# GAN-based Algorithm for Efficient Image Inpainting.

Zhengyang Han[a†], Zehao Jiang[b†], Yuan Ju[*c†]
[a]New York University, College of Arts and Science, New York, United States
[b]Georgia Institute of Technology, College of Science, Atlanta, Georgia, United States
[c]Rensselear Polytechnic Institute, School of Science, Albany, New York, United States
[*] Corresponding author: juy@rpi.edu
[†] All these authors are equally contributed.

## ABSTRACT

Global pandemic due to the spread of COVID-19 has post challenges in a new dimension on facial recognition, where people start to wear masks. Under such condition, the authors consider utilizing machine learning in image inpainting to tackle the problem, by complete the possible face that is originally covered in mask. In particular, autoencoder has great potential on retaining important, general features of the image as well as the generative power of the generative adversarial network (GAN). The authors implement a combination of the two models, context encoders and explain how it combines the power of the two models and train the model with 50,000 images of influencers faces and yields a solid result that still contains space for improvements. Furthermore, the authors discuss some shortcomings with the model, their possible improvements, as well as some area of study for future investigation for applicative perspective, as well as directions to further enhance and refine the model.

## 1. INTRODUCTION

Generative Adversarial Network (GAN) is First introduced by American Computer Scientist Ian Goodfellow in 2014, which has seen become a popular generative method in the area of machine learning [1]. GAN requires two separately trained models, the discriminator that classifies the generated data from the real data set, and the data generator to generate data similar to the real ones in order to fool the trained discriminator. This model is widely used to generate images of different styles, to enhance image resolution, or image blending, etc. It is also capable of image inpainting, completing a partially masked or blurred image.

In fact, as a generative method, GAN has demonstrated solid performance when tackling a specific task, i.e., image inpainting. Image inpainting is the process to fill a damaged image, with a particular region either blackout or covered, with coherent content in relation to the rest of the images. In particular, facial inpainting is the task of image inpainting where the damaged region is a part of a human's face. During the covid era, when more populations endorse masks to prevent the spread of disease, the aspect of face recognition faces new challenges. Wearing masks poses new challenges to the already mature face recognition tasks, as masks cover important facial features and details for the computer to recognize a person. Essentially, images or photos of people wearing masks work the same way as those damaged or flawed face photos, and one solution is face inpainting, reconstructing the original faces without losing too much detail. As such, scientists and machine learning engineers are looking at options to utilize GAN in masked face recognition. Date back to 2017, a group of computer vision researchers proposed the generative model in face completion [1]. This group constructs a GAN model where the discriminator relies on multiple different loss function, which accounts for reconstruction loss, adversarial loss, and semantic loss and yields results that reflect the performance quantitatively and qualitatively.

In 2020, a group of Chinese scientists, inspired by the effectiveness of recurrent neural networks (RNN) in low-level image tasks and the generative power of GAN, proposed ideas to construct a recurrent generative adversarial network for face inpainting [2]. This group of scientists constructed a recurrent GAN, in which the generator consists of two convolutional neural networks (CNN), and a RNN, which effectively utilizes any possible relationship between different captured features. According to the scientists, their model outperformed other models, including GIICA and GLCIC, in completing facial images. During the same year, a group of Korean computer engineers proposed a novel strategy to tackle the task [3]. They trained two separate discriminators, one checks the inpainting part in detail, the other assures global coherency. The generator was then trained to simultaneously fool both discriminators. Their model was also reported to perform with reliable results when tackling masked faces or damaged images in comparison to other state-of-the-art image inpainting methods.

Around the same time, another group of Chinese scientists considered the possibility of utilizing the Coherent Semantic Attention (CSA) layer in the regular GAN structure [4]. This layer emphasizes the correlation between the generated portion of the image to the context of the image by searching through the rest of the image when training the generator. Those scientists also introduce the concept of consistency loss in addition to the regular loss function, which allows the model to adjust its CSA layer. Their model is, in fact, also effective in generating specific missing regions when tackling image inpainting task. In addition, similar approaches are also endorsed by researchers on contextual attention when tackling generative image inpainting [5].

Context encoder, on the other hand, is a model consisting of a generator based on autoencoder and an adversarial discriminator that achieves image inpainting [6]. The model works in a similar way to GAN, only that its generator, a denoising autoencoder consisting of multiple layers of encoders and decoders. This autoencoder is trained in a similar way to other autoencoders, in an unsupervised learning way. Each encoder layer preserves more general and more abstract features, and when decoding layers attempt to reconstruct the image, the generator forgets about the masks or flaws and generates images only based on learnt features. Hence, this model is able to fulfill damaged images while learning to fool the adversarial discriminator.

## 2. OUR METHOD AND MODEL

### 2.1 Overview

In this paper, the authors aim to implement a baseline GAN model consisting of a generic discriminator, and then implement a generator consisting of layers of encoders and decoders. The authors then examine the possible shortcomings of the model and attempt to refine the resolution, clarity, and convergence speed of our model. The authors also attempt to enhance the model in an applicative aspect to achieve face inpainting tasks. The authors construct a context encoder based on the general structure introduced by Pathak and implement it, rather on street view, but on facial images. The authors then examine the performance of the entire model through different lenses and discuss possible further improvements in either algorithms or possible applications and future research aspects. The framework is shown in Figure 1.

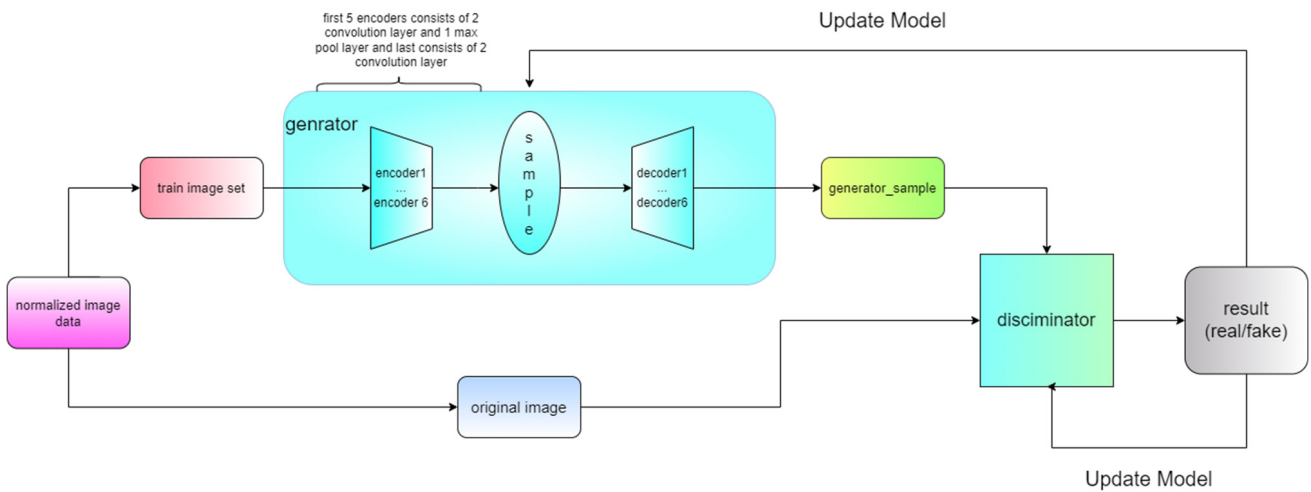

Figure 1. Framework of our utilized method.

### 2.2 Discriminator

This adversarial discriminator is designed to distinguish between the real image data provided by the dataset from the generated image data from the generator of this network structure, as in most GAN. To enhance the equality of the generated images and widen the possible areas where our model can be applied to, the authors have altered our model from a generic GAN to a context encoder. This model is designed to deal with image of size 512✕512. The discriminator, instead of 5 keras dense layers, are consisted of 6 sub-discriminator unit fully connected with each other, each unit consists of 2 convolution layers, with kernel size of 3✕3 and 2✕2, respective. Each unit, except for the last unit, also contains a max pooling layer at the end to reduce the invariance and preserve dominant features as opposed to minor details. All hidden layers endorse the exponential linear unit (ELU) as an activation function, while the output layer uses sigmoid activation function.

### 2.3 Generator

This generator is designed to learn to generate images close to the original ones provided by the dataset with sufficient coherence to challenge the discriminator's ability to distinguish between the real ones and the generated ones, as in a normal GAN. However, this generator is not an usual convolutional neural network, but rather an autoencoder consists of an encoder and a decoder. This particular structure allows the encoder to process the original incomplete or damaged images and only encode the abstract or general features of the images when it reaches an information bottleneck. Hence, as the decoder reconstructs, the decoder forgets about the missing part and generates the image with coherency to the rest of the image. The encoder has a similar structure to the discriminator, with 6 sub-encoder units, each with 2 convolution layers and, except for the last unit, followed by a max pooling layer to reduce variance or other influences from any rotations or shifts. Then, the three channels, each representing one color of RGB, is reshaped to be stored as a latent vector. The encoder is then followed by a decoder consisting of 6 sub-decoder units. The size of the decoder layers is the same as those in the encoders but in reverse order, since its aim is to reconstruct the image. Each sub-decoder unit is consisting of 2 convolution layers, with kernel size 3✕3 and 2✕2 respectively. There is no pooling layer necessary for the decoder, and the activation function is ELU.

### 2.4 Loss

The loss for this model consists of two different losses, the reconstruction loss and the adversarial loss, which is shown in Eq.(1). The reconstruction loss of the autoencoder generator is l2 loss that captures the feature of missing or damaged region as well as the context of the image. The adversarial loss, on the other hand, is based on the negative logistic likelihood when the discriminator attempts to distinguish the images. When training the generator, however, the authors utilize a total loss, combining both adversarial loss from the discriminator and the reconstruction loss. This total loss preserves both the quality of the reconstructed image from the adversarial loss and the context of the image from the reconstruction loss. For each epoch, the overall loss is the mean square error (MSE) of all training or testing images.

$$L_{total} = L_{reconstruction} + L_{adversarial} \quad (1)$$

### 2.5 Implementation details

With the selected dataset loaded and normalized properly, the authors then train the model on a local computer with downloaded celeb-A data. The minimum batch size is set to 8 at default and the model will be trained for 20 epochs at default. The learning rate is set at 1e-4. Further investigation on this learning rate and possible overfitting or underfitting is done later. The authors use Adam optimizer to allow the model to converge during training and first train the adversarial discriminator and then the generator. See result section for the results of this trained model.

## 3. EXPERIMENTS

### 3.1 Dataset description

The dataset used for training and testing the model is the Influencer Human Real Face Data Set provided by Set Pretty Face, with resolution 128x128. This model is also capable of learning other featured image dataset, including but not limited to street views, mountain views, animated characters, vehicles, etc. The number of images chosen for the whole training and testing set is about 50000 images and split it in to 0.9:0.11 for Training: Testing.

### 3.2 Comparison results

The learning curve is shown in Figure 2. After training 20 epochs (each with 704 steps) for 14080 iterations with batch size equals to 64, our method achieves a good convergence, within a training time of 10 hours, 8 minutes, and 58 seconds. The total loss decreases a large amount at the first 1000 iterations, and the rate of decrease gradually becomes low and when the number of iterations larger than 10000, the value of total loss oscillates around 70. The results are listed in Figure 3. The result indicates solid potential in our model on image inpainting.

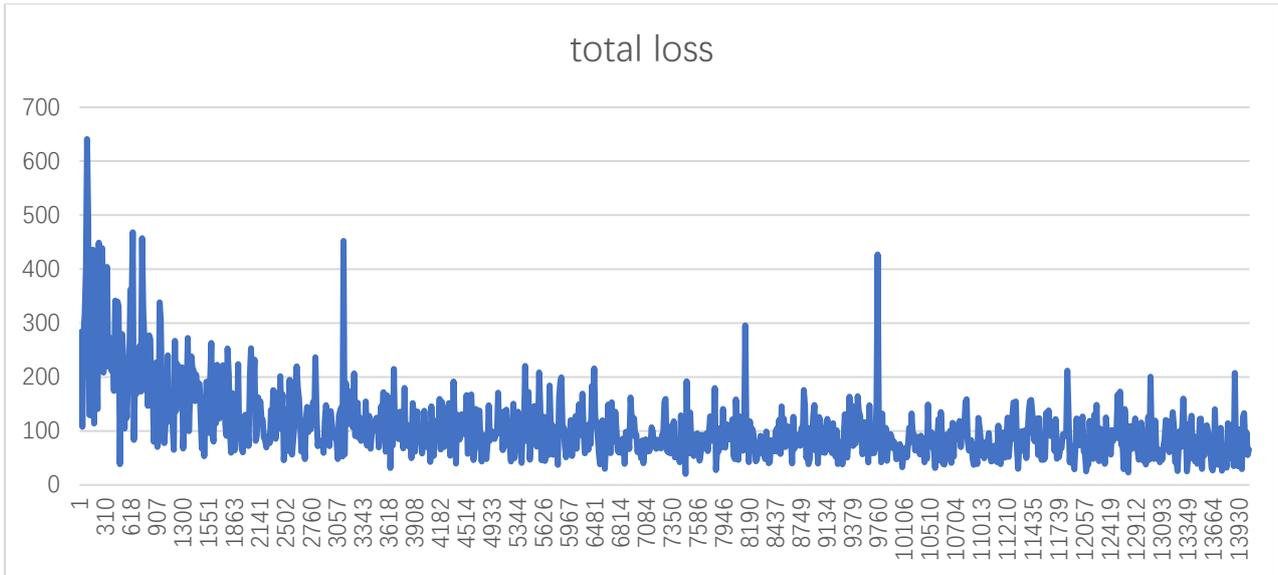

Figure 2. Training loss curve during training.

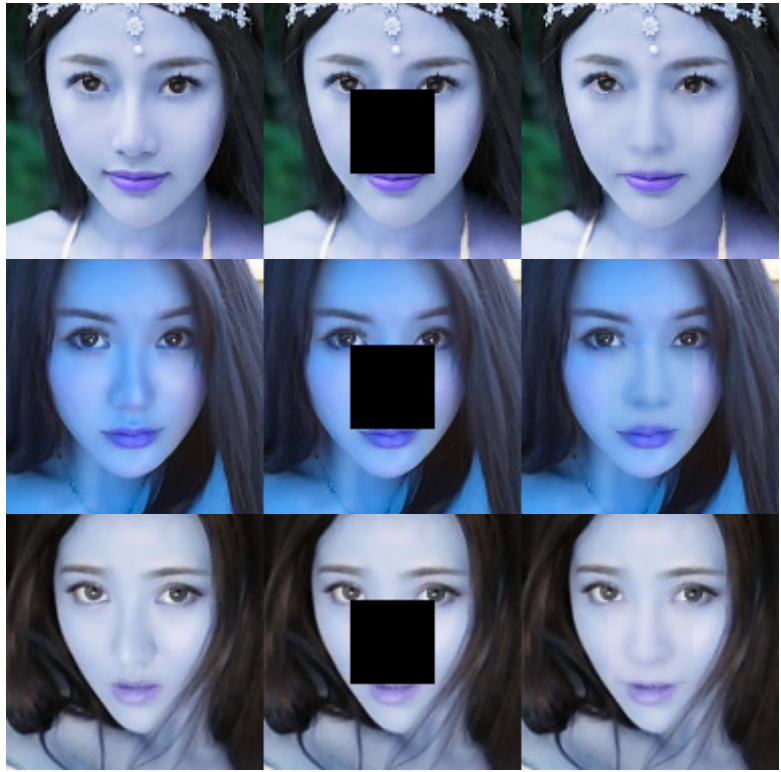

Figure 3. Sample testing output

### 3.3 Discussion

Our model leaves space for improvements. Here are some issues the authors encounter during the project and a possible improvement that the authors consider implementing in future research and investigation. One issue, experienced in both the original model by Pathak and in our model, is the extensive time taken to train the model. While the authors do not record the exact training time, it takes approximately 10 hours to complete 20 epochs with a batch size of 64. While it seems an acceptable on the time scale, consider the fact the authors are training based on image size 128✕128, training datasets with larger image size, such as 512×512 for example, may take longer time, possibly on an exponential scale. Hence, improvements can be made on the shortening the training time. It is possible that the model requires less training time once given a proper batch size. Using GPU instead of CPU should also allow the model to converge faster. Most importantly, however, the authors propose a possible improvement: adding batch normalization layers. The addition of batch normalization layers allows the output of hidden layers to be normalized before passing to the next corresponding layers. It ensures that all possible features in an output of a layer are on the same scale [7,8]. Hence, when hidden layers attempt to capture important features, batch normalization layers are expected to avoid larger updates on one weight, small ones on another at the same time, and thus, the model converges faster.

Furthermore, the authors state that image inpainting can be a potential solution to recognizing a human's masked face. Consider the possibility to first implement a CNN-based object detector, or an object removal network specifically trained to detect masks. This object detector should return the exact pixels where the mask is in the image, and then those returned pixels can be either blurred or blacked out. In 2022, researcher Yuan Zhou has implemented such face detector that relies on mobile net, which the authors believe can be further explored and refined [9]. Our context encoder now completes the partially blurred or blacked out image. Hence, our model can potentially be applied to, for example, face ID checking, where with the context encoder, the face recognition discriminator now only requires checking the reconstructed images of people without masks. Depending on the training dataset, the model can also be applied to remove the effects of rain or fog for photographers or even for surveillance cameras.

As discussed, scientists are considering the possibility of using a recurrent GAN to tackle image completion tasks. Unlike regular neural networks, RNN retains some memory of the sequential features of the data. This characteristic may allow RNN to outperform other models in generative images or data where the surrounding or context is crucial. the authors hence suggest that further investigation can be focused on how to implement RNN rather than regular CNN in both generator and discriminator, and how such new model, recurrent GAN performs in comparison to the regular GAN or context encoder in the area of image inpainting.

Apart from RGAN, DRGAN, disentangled representation learning GAN can also be a possible direction for face inpainting [10]. In 2017, researchers have already started to investigate the possibility of using DRGAN to tackle a serious problem in face recognition, that most images captured in real life feature different poses. DRGAN allows the model to learn and adjust different variations of people's face and body. The combination of DRGAN and context encoder may have the potential to solve complex face inpainting problems and is worth further consideration.

## 4. CONCLUSION

In this paper, the authors aim to exploit the effectiveness of GAN models on image inpainting task. The authors design a model, which consists of a generator, a discriminator. Our context encoder model takes approximately 21 hours to complete the default 20 epochs of training at a local computer. The training and testing are done with a dataset of 2500 images of the 50,000 entire Influencer Human Real Face Dataset. The visualization results validate that the GAN model can handle the image inpainting task with satisfying performance. Furthermore, the authors also listed our limitations and future work for other researchers to design a much better GAN based image inpainting models.

## REFERENCES

[1] Dhamecha. (2021). A Detailed Explanation of GAN with Implementation Using Tensorflow and Keras. Data Science Blogathon.
[2] Wang, Q., Fan, H., Sun, G., Ren, W., & Tang, Y. (2020). Recurrent generative adversarial network for face completion. IEEE Transactions on Multimedia, 23, 429-442.
[3] Din, N. U., Javed, K., Bae, S., & Yi, J. (2020). A novel GAN-based network for unmasking of masked face. IEEE Access, 8, 44276-44287.
[4] Liu, H., Jiang, B., Xiao, Y., & Yang, C. (2019). Coherent semantic attention for image inpainting. In *Proceedings of the IEEE/CVF International Conference on Computer Vision* (pp. 4170-4179).
[5] Yu, J., Lin, Z., Yang, J., Shen, X., Lu, X., & Huang, T. S. (2018). Generative image inpainting with contextual attention. In *Proceedings of the IEEE conference on computer vision and pattern recognition* (pp. 5505-5514).

[6] Pathak, D., Krahenbuhl, P., Donahue, J., Darrell, T., & Efros, A. A. (2016). Context encoders: Feature learning by inpainting. In Proceedings of the IEEE conference on computer vision and pattern recognition (pp. 2536-2544).

[7] Chen, L., Fei, H., Xiao, Y., He, J., & Li, H. (2017, July). Why batch normalization works? a buckling perspective. In *2017 IEEE International Conference on Information and Automation (ICIA)* (pp. 1184-1189). IEEE.

[8] Wu, S., Li, G., Deng, L., Liu, L., Wu, D., Xie, Y., & Shi, L. (2018). $L1$-norm batch normalization for efficient training of deep neural networks. *IEEE transactions on neural networks and learning systems*, *30*(7), 2043-2051.

[9] Zhou, Y. (2022). The Efficient Implementation of Face Mask Detection Using MobileNet. In Journal of Physics: Conference Series (Vol. 2181, No. 1, p. 012022). IOP Publishing.

[10] Tran, L., Yin, X., & Liu, X. (2017). Disentangled representation learning gan for pose-invariant face recognition. In *Proceedings of the IEEE conference on computer vision and pattern recognition* (pp. 1415-1424).